# Saccade Attention Networks: Using Transfer Learning of Attention to Reduce Network Sizes

Marc Estafanous

[estafanous@neurobaby.com](mailto:estafanous@neurobaby.com) / mestafa1@jhu.edu

**Abstract :**

One of the limitations of transformer networks is the sequence length due to the quadratic nature of the attention matrix. Classical self attention uses the entire sequence length, however, the actual attention being used is sparse. Humans use a form of sparse attention when analyzing an image or scene called saccades. Focusing on key features greatly reduces computation time. By using a network (Saccade Attention Network) to learn where to attend from a large pre-trained model, we can use it to pre-process images and greatly reduce network size by reducing the input sequence length to just the key features being attended to. Our results indicate that you can reduce calculations by close to 80% and produce similar results.

**Introduction:**

In this project, we introduce Saccade Attention Networks. We will use pretrained vision attention networks (teacher network) to provide the targets for saccade like attention in second networks (student networks) that can then be smaller and hopefully have similar results.

Evolution has had millions of years to experiment with vision processing. Nearly all animals with eyes use saccade like movements to focus their attention on objects (Land MF , 1999). Humans are of course included in this cohort, however, we are not born with the ability to saccade efficiently but in fact learn how to over time (Alahyane, 2022). Once they are learned, humans are able to classify images faster than with random movements. Following human saccades has demonstrated that that saccades are preplanned prior to the movement and based on predicted locations of semantic features (Land, 1999).

**Related Work**

Several neural networks have been designed to mimic these movements. The methods use various techniques to learn to predict the saccade locations including RNN (Xu, 2015), hard attention (Elsayed, 2019), gradient descent (Tan, 2021), and reinforcement learning (Huang, 2022). The latter article showed that by using saccade like glances, networks can be more efficient.

In each case, the saccades are learned from scratch during the training of the network. The hypothesis we explore here is that using transfer learning of the points of attention from pretrained vision transformer networks to learn where to saccade, we can train more efficient networks and achieve similar results as the teacher network.

If the hypothesis is correct, there is the potential that neural network models used in vision processing (and possibly later auditory, somatic and other sensory processing), could have their results reproduced with smaller, more efficient networks.

We will be using transfer attention learning to introduce a new form of attention, Saccade Attention, which is used in Saccade Attention Networks, and test the performance of these networks on the CIFAR100 dataset. The performances will be judged by the overall accuracies after a reasonable number of iterations.

Contributions of this work:

1) We develop a new method of transfer learning that transfers attention from large transformer networks to much smaller ones using biologically inspired saccade-like attention.
2) We show that while the method as a whole may not work, it is possible to learn ViT attended patches with a CNN.

**Methods:**

In the original transformer paper, (Vaswani, 2017), the use of self attention and positional embeddings changed the world that we live in by creating an effective learning unit they called a transformer.

Self attention is used to determine which elements in a sequence are important to focus on at any given time. In one sense, images are sequences just like text, so that realization led to the development of vision transformers (ViT) (Dosovitskiy, 2021).

While it is difficult to gather human saccade information, and it is possible to learn saccadic motions from scratch, the nature of the attention mechanism in the transformers seems lend itself to be a valid source of vast amounts of data with regards to where a network may want to saccade to. Given that saccade networks can be more efficient (Huang, 2022) than regular networks, it is natural to conclude that using networks that have already learned where to attend to train smaller more efficient saccade networks may be possible.

Furthermore, the training can use transformer networks to learn since the saccade location sets are simply sequences, so reinforcement learning should not be necessary.

The first step we need is to select a teacher model from which to extract the most attended to patches for any image. The teacher network can be any Vit, but we will use a pretrained Dino model from Facebook. This model was chosen based on visualization of the attention maps which appear to show the most clearly defined areas of attention (Fig. 1, first row) with larger patch sizes. Smaller patch sizes of course showed greater resolution of the attention but would entail greater numbers of calculations. We chose a model with 16x16 patches to keep network size smaller.

The attention was extracted from the query-key values of the model using a method called attention rollout (Abnar, et. al., 2020). The idea behind attention rollout is that the attention can be deduced from the CLS token which represents the entire image since it is the first token in a transformer. The attention is calculating by multiplying the attention matrix in each layer but carrying forward the previous attention by adding an identity matrix and then normalizing the results. The attention for the CLS token is then sliced out and a heat map is created (Fig 1).

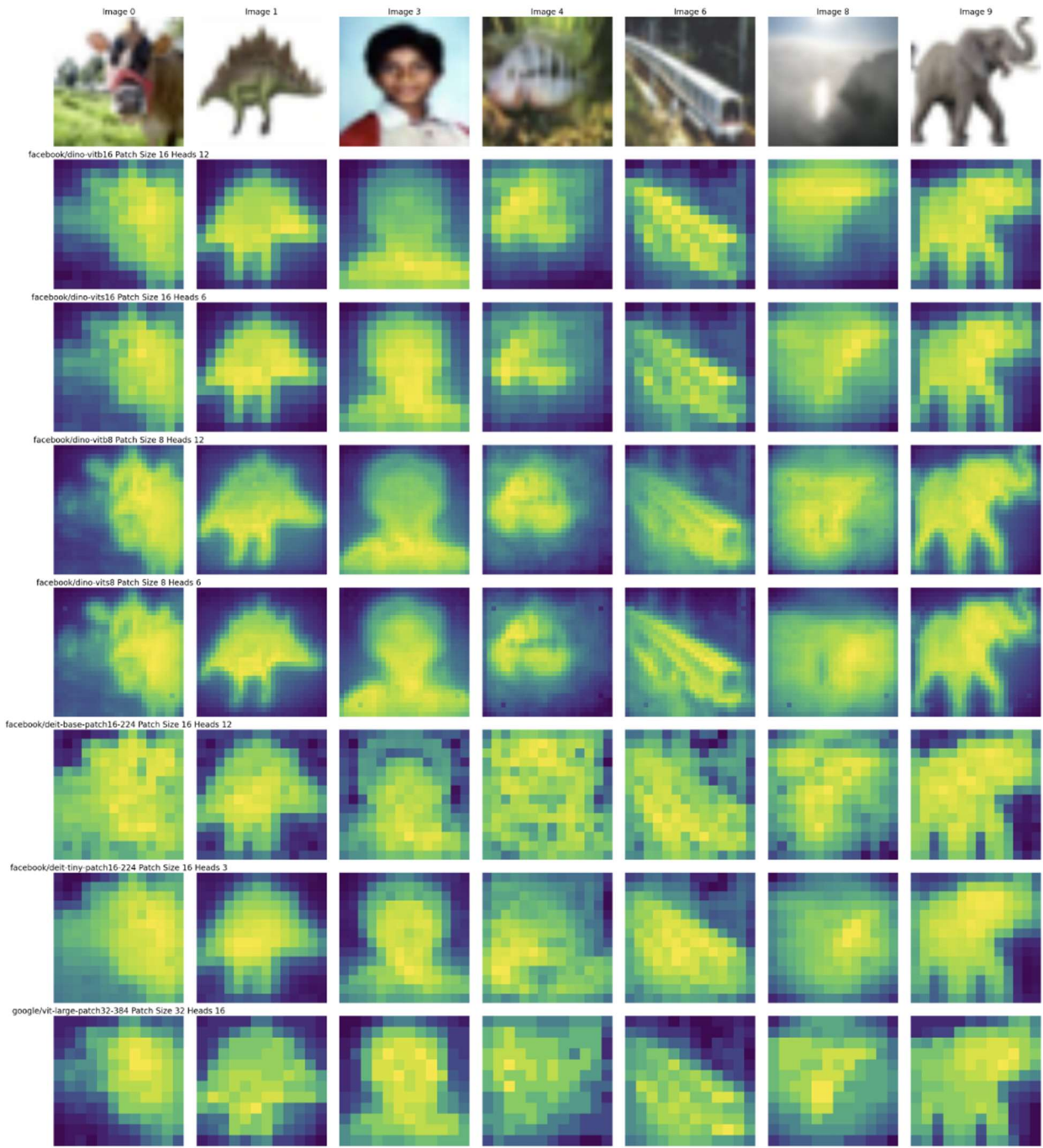

*Figure 1: Attention maps of various ViTs*

The second step is to extract the top-k patches of attention. The indices of these patches are stored as the target of a custom dataset that consists of the original image and the indices. The dataset is then saved for future use. It is not surprising to note that most of the attended to patches are near the middle of the images (Figs 2-3).

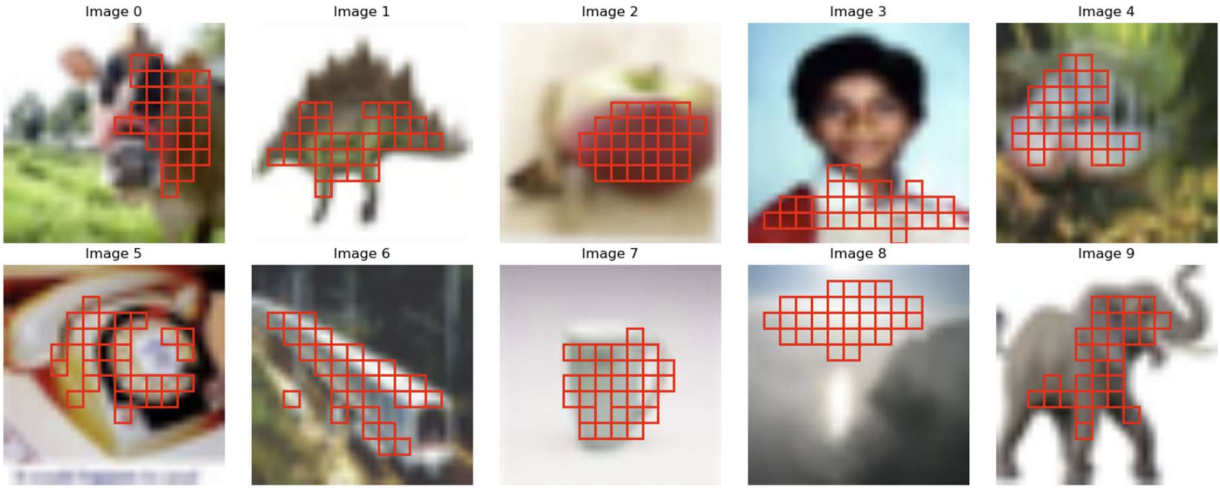

Figure 2: Top-K patches over images

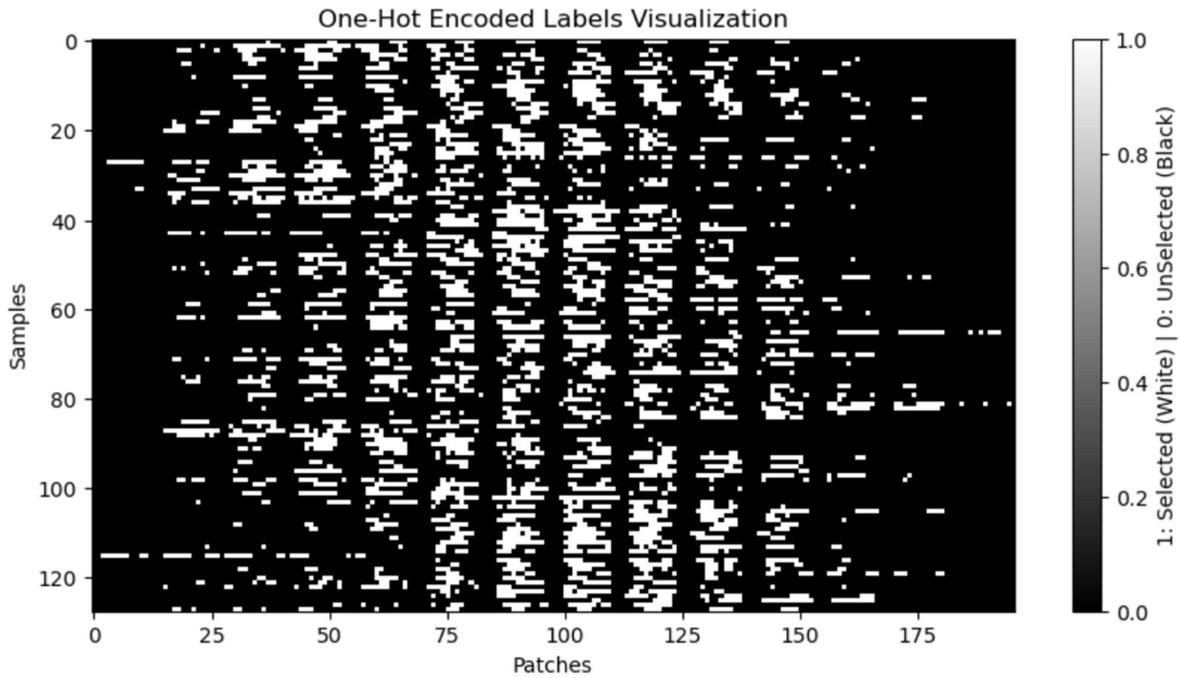

Figure 3: Top-K indices in white of various samples.

Next, a classifier network is trained to learn the multi-class output of the indices based on an embedding of the image. The network we chose to use is a pretrained Resnet18 model. The classifier layers were removed and replaced with 196 outputs (1 for each patch) that represented the patch indices. The network is fine-tuned using a BCE loss to output the patch indices to be paid attention to by the student network. This teacher network is called a Saccade Attention Network (SAN).

Finally, a student network, which is a ViT, is trained to learn the images and is called a Saccade Attention Network Vision Transformer (SAN-Vit). During both training and testing, an image is first fed into the SAN and the predicted indices for attention are determined. The patches are then sliced out of the whole set of patches, allowing for maintenance of the gradient. The sliced patches are then attended to with the self attention of the SAN-Vit and classified.

The size of the self attention in the ViT is reduced from the original seq_len * seq_len to k * k. If there are 16 x 16 patches for example in a 224 x 224 image, then we have 14x14 patches to attend to. That's 196 x 196, or 38,416 comparisons. We use k=32 which limits the attention to only 32 x 32, or just 1,024 comparisons, a 38x reduction.

Data augmentation included moving and resizing the images with torch RandomSizedCrop, as well as horizontal flipping.

We will use one of the original ViT models from Facebook which is pretrained on ImageNet-1k, and is available on hugging face (facebook/dino-vitb16). We will test the model on CIFAR-100. More datasets may be compared in the future. Early stopping was used when the validation set results stopped improving. Multiple variations of the model was tried including various sizes, and positional embedding types. Current CIFAR scores are in the 80-90% range and Dino scored 91% when fine-tuned.

The metrics used will be overall accuracy for all datasets as well as the network sizes and estimated FLOPs (FLOating Point operations). FLOPs are the number of calculations performed by the network in a forward pass. The current network results are not impressive. We will need to compare 3 networks in the final analysis:

1) The original Dino ViT fine-tuned on the dataset.
2) An equivalent sized ViT but with the full attention matrix
3) The SAN Vit

**Results:**

The model reduces the attention calculations by 7x using the parameters above compared to a simple ViT. Overall, there is less of a reduction than just the attention calculation savings since we're adding the overhead of a ResNet classifier, but overall the calculations are still significantly reduced. In fact, there is potentially an approximately 79% reduction in the number of calculations with only slightly more parameters compared to a simple ViT model (Table 1). Both models are significantly leaner compared to Dino, which can explain the results with regards to accuracy discussed below.

*Table 1: Estimated Parameter and Flop Counts (source: chatGPT)*

| Model | Patch Count | Transformer Dim | Depth | Params | FLOPs | Notes |
| --- | --- | --- | --- | --- | --- | --- |
| facebook/dino-vitb16 | 196 | 768 | 12 | 85.9M | 17.36 GFLOPs | From `torchinfo`; DINO ViT-Base |
| Simple ViT | 196 | 128 | 4 | 2.5M | ~0.19 GFLOPs | Small custom transformer |
| ResNet-18 + SAT | 32 | 128 | 4 | 14.8M | ~0.04 GFLOPs | ResNet backbone + lightweight ViT |

First, let's see how well the SAN faired. It performed well and achieved 84% on the test set for choosing the correct patches to attend to, with similar sensitivity and specificity. We can see how well the network does by visualizing the patch selection in Figures 4 and 5.

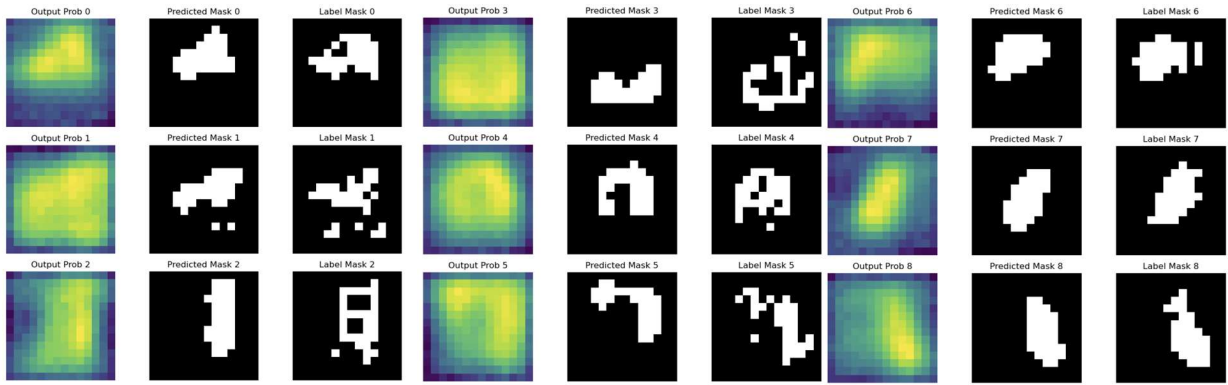

*Figure 4: SAN Attended Patches vs Ground Truth*

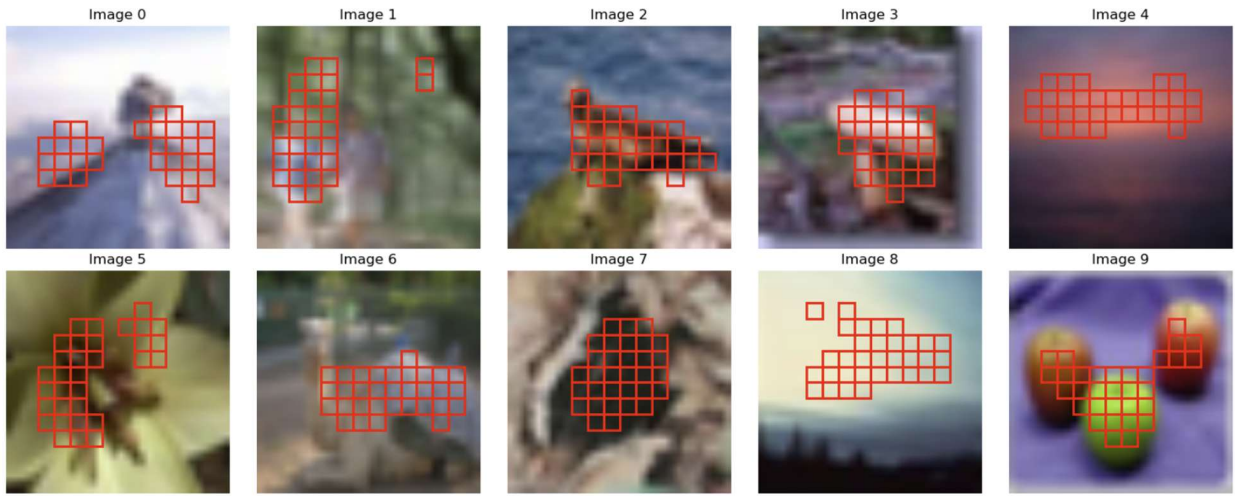

*Figure 5: SAN Attended Patches*

As we can see from the images, the attended patches do a good job of focusing on the images' main objects.

When we used those patches in to the SAN-Vit, the results were unfortunately very poor (Table 2). When trained for 100 iterations, the training set over fit and achieved a 95% accuracy, however the test and validation sets never scored much higher than 30%.

*Table 2: Model Results*

| Model | Test Accuracy |
|---|---|
| DINO-ViT-B16 | 62.58% |
| Simple-ViT | 31.60% |
| SAN-ViT | 30.23% |
| SAN-GT | 26.78% |
| SAN-GT-Set | 24.06% |

Without data augmentation, the results never beat 25%. Using a non-sinusoidal positional embedding on the full image did not change the results. Using a learned positional embedding after the patch slicing had little effect on the results. Using a larger model reduced the results slightly due to more overfitting. Removing positional embeddings altogether reduced the results.

In order to determine where the network is failing, we first looked at using the ground truth labels instead of the SAN labels (SAN-GT). The results did not improve and actually worsened to 26%. When trying to remove positional embeddings to see if a set of image patches was better, the results worsened further to 24% (SAN-GT-Set).

It's possible, the issue is the size of the network. Let's look at a ViT model that uses all the patches, i.e. the full 196 patches. This model failed as well with only 31% accuracy. This is interesting though because the SAN-ViT Scored the same with only 21% of the FLOPs. In a sense, this means the model worked compared to a standard, simple ViT: we got the same accuracy with significantly fewer calculations.

In order to see if it's because of the dataset size, let's see how a pretrained model does and just fine tune the original Facebook Dino model. This does way better and scored more than double the other models' scores with 64% accuracy. This could be for one of two reasons. One, the model is way bigger, and two, it was pretrained on ImageNet-1k. It is unclear though why it maxed out at 64% accuracy versus the reported 90%, but may have to do with fewer data augmentations. Each training iteration took well over an hour though, so it was not possible to do further testing.

**Conclusions:**

Preliminary runs of the network show mixed results. The SAN did well at predicting where to pay attention to. The SAN-Vit however did nowhere near as well as the state of the art and less than half as well as the fine-tuned Dino model on the same data.

One other interesting thing to note is that when Facebook AI Research published their Masked Autoencoder paper (He, et. al., 2021), they found that the ideal number of patches to use is in the -30-118 range (Figure 5 of their paper). It is possible that increasing the number of patches may increase the accuracy, but the difference in percent performance was less than 1% throughout that entire range of patches on Imagenet-1k. What that paper does tell us though is that it should be possible to perform much better with the selected patches than what we're seeing. If it's not due to network size or data augmentation, then the issue is with the network architecture as a whole.

In theory, the method could be applied to any network that used transformers to learn, but first many more studies need to be performed if more compute or time becomes available. Specifically, we would try larger models, e.g. Dino but with fewer patches, running a blank Dino that's not fine-tuning to compare results without pre-training, larger simple Vits, more data augmentation, and more datasets. We would first need to reproduce the 90% performance on CIFAR-100 and then adjust the experimental models accordingly. Future work would also include using a transformer network for the saccade attention and trying on text corpuses.

If this works, the method could significantly reduce the size and increase the efficiency of any transformer networks.